%% file: camera_ready.tex
\documentclass[letterpaper]{article} 
\usepackage{aaai2026}  

\usepackage{times}  
\usepackage{helvet}  
\usepackage{courier}  
\usepackage[hyphens]{url}  
\usepackage{graphicx} 
\usepackage{natbib}  
\usepackage{caption} 
\usepackage{algorithm}
\usepackage{algorithmic}
\usepackage{xcolor}

\usepackage{amsmath}
\usepackage{amsthm}
\usepackage{amssymb}
\usepackage{multirow} 
\usepackage{booktabs}
\usepackage{pifont}

\usepackage{newfloat}
\usepackage{listings}
\DeclareCaptionStyle{ruled}{labelfont=normalfont,labelsep=colon,strut=off} 
\floatstyle{ruled}
\newfloat{listing}{tb}{lst}{}
\floatname{listing}{Listing}
\title{Open-World 3D Scene Graph Generation for Retrieval-Augmented Reasoning}
\author{
    Yu Fei\textsuperscript{\rm 1}, Quan Deng\textsuperscript{\rm 2}, Shengeng Tang\textsuperscript{\rm 3},  Yuehua Li\textsuperscript{\rm 4}, Lechao Cheng\textsuperscript{\rm 3}\thanks{Correspondding Author: chenglc@hfut.edu.cn}\\
}
\affiliations{
    \textsuperscript{\rm 1} Liaoning University of Technology\\
    \textsuperscript{\rm 2} University of the Chinese Academy of Sciences\\

    \textsuperscript{\rm 3} Hefei University of Technology\\

    \textsuperscript{\rm 4} Zhejiang Lab\\

}

\usepackage{bibentry}

\begin{document}

\maketitle

\input{Sections/0_abstract}

\input{Sections/1_introduction}

\input{Sections/2_related_works}
\input{Sections/3_method}

\input{Sections/4_experiments}
\input{Sections/5_conclusion}

\bigskip

\bibliography{aaai2026}
\clearpage
\input{supplements-sections/0_Appendix}
\input{supplements-sections/1_task1}

\input{supplements-sections/2_task2}

\input{supplements-sections/3_task3}

\input{supplements-sections/4_task4}

\end{document}

%% file: Sections/0_abstract.tex
\begin{abstract}
Understanding 3D scenes in open-world settings poses fundamental challenges for vision and robotics, particularly due to the limitations of closed-vocabulary supervision and static annotations. To address this, we propose a unified framework for Open-World 3D Scene Graph Generation with Retrieval-Augmented Reasoning, which enables generalizable and interactive 3D scene understanding. Our method integrates Vision-Language Models (VLMs) with retrieval-based reasoning to support multimodal exploration and language-guided interaction. The framework comprises two key components: (1) a dynamic scene graph generation module that detects objects and infers semantic relationships without fixed label sets, and (2) a retrieval-augmented reasoning pipeline that encodes scene graphs into a vector database to support text/image-conditioned queries. We evaluate our method on 3DSSG and Replica benchmarks across four tasks—scene question answering, visual grounding, instance retrieval, and task planning—demonstrating robust generalization and superior performance in diverse environments. Our results highlight the effectiveness of combining open-vocabulary perception with retrieval-based reasoning for scalable 3D scene understanding.

\end{abstract}

%% file: Sections/1_introduction.tex
\section{Introduction}

Understanding 3D scenes is fundamental for tasks like autonomous navigation and augmented reality. However, most traditional methods rely on closed-vocabulary annotations and curated datasets, limiting their scalability in dynamic, unstructured environments where novel objects and interactions frequently emerge. This gap calls for open-world 3D understanding methods that can generalize to unseen scenarios without exhaustive supervision.

Recent advances in vision-language models (VLMs) such as CLIP~\cite{hafner2021clip}, ALIGN~\cite{jia2021scaling}, and ImageBind~\cite{girdhar2023imagebind} have enabled open-vocabulary learning, which shows remarkable potential in 2D tasks like classification and detection. These successes have inspired efforts to adapt VLMs for 3D vision, leading to progress in tasks such as semantic segmentation and object recognition. However, these methods often rely on 2D-3D projections and RGB-D inputs with known poses, which are impractical in many real-world settings due to sensor limitations, occlusions, and viewpoint variations.

To address these limitations, we introduce a retrieval-augmented open-world 3D scene graph generation framework. It consists of two core components: (1) a dynamic scene graph module that constructs semantic and spatial representations of the environment by identifying objects and relationships without fixed label sets; and (2) a retrieval-augmented navigation system that enables natural language-based exploration of 3D scenes. 



This work is motivated by the need for adaptive and scalable 3D understanding systems that can operate in open-world settings without extensive human supervision. While conventional methods perform well in constrained environments, they struggle to generalize to dynamic and diverse scenarios. By integrating VLMs with retrieval mechanisms, our method enables compositional reasoning over unseen objects and relationships. Our framework addresses the following challenges: (1) \textbf{Open-Vocabulary Grounding}: Recognizing and localizing unseen objects and relations without predefined labels. (2) \textbf{Dynamic Graph Construction}: Building and updating 3D scene graphs that capture evolving semantic and spatial relationships. (3) \textbf{Language-Guided Interaction}: Supporting navigation and querying through natural language, requiring alignment between textual and spatial modalities.

Our contributions are summarized as follows:
\begin{itemize}
    \item A novel framework for open-world 3D scene graph generation, leveraging vision-language models to extract objects and relationships without fixed annotations.
    \item A retrieval-augmented reasoning module that enables flexible scene interaction through query-based exploration.
    \item Extensive experiments on 3DSSG and Replica benchmarks demonstrating improved performance and generalization in dynamic, real-world environments.
\end{itemize}

%% file: Sections/2_related_works.tex
\section{Related Work}
\label{sec:related}

\textbf{Open-Vocabulary Scene Understanding.} 
The success of 2D vision-language models (VLMs) such as CLIP~\cite{hafner2021clip}, ALIGN~\cite{jia2021scaling}, and ImageBind~\cite{girdhar2023imagebind} has spurred interest in extending open-vocabulary capabilities~\cite{wang2024marvelovd, tian2024open} to 3D tasks, including semantic and instance segmentation~\cite{ha2022semantic, hegde2023clip, zhang2022pointclip}. These early efforts often fuse CLIP-based supervision with 3D backbones using RGB-D data with known poses, but their performance is limited by projection errors, occlusion artifacts, and strong reliance on pose priors.

\noindent  \textbf{Open-Vocabulary Scene Graph Generation.}
3D scene graph generation aims to model both semantic and spatial relationships among objects, serving as a bridge between visual perception and symbolic reasoning. Prior work such as~\cite{wald2020learning, ren2023sg} focused on structured indoor data with supervised labels. Recent methods like Open3DSG~\cite{koch2024open3dsg} and zero-shot variants~\cite{linok2024beyond} leverage VLMs to recognize unseen objects and infer spatial relations. However, many still depend on annotated RGB-D inputs or fixed camera poses, which limits their adaptability to open-world settings.
In contrast, our method eliminates the need for manual annotations or fixed-pose RGB-D data, enabling fully annotation-free 3D scene graph generation. It achieves strong performance in both open- and closed-vocabulary benchmarks while offering better generalization to unseen environments.

\noindent \textbf{Efficient Multimodal Large Language Models Adaptation.} 
Multimodal Large Language Models (MLLMs) have emerged as a powerful paradigm for enabling joint reasoning over visual and textual inputs~\cite{wu2023survey, hu2023cell, che2024multimodality, wang2024entityclip, zhang2025asap, wang2025beyond, li2025temporal, shen2025beyond}. These models typically comprise three components: a large language model (LLM) backbone (e.g., LLaMA~\cite{dubey2024llama}, Qwen~\cite{bai2023qwen}), a visual encoder (e.g., CLIP~\cite{Radford2021clip}, BLIP~\cite{Li2022blip}), and adapter modules that align visual features with the LLM’s embedding space~\cite{yin2023adapter}. This architecture supports a wide range of vision-language tasks, such as image captioning, visual question answering, and visual grounding. Given the size and complexity of MLLMs, training them from scratch is computationally expensive~\cite{jiang2025dcp}. To address this, Parameter-Efficient Fine-Tuning (PEFT) methods~\cite{Houlsby2019peft}—such as LoRA~\cite{hu2022lora} and Prefix Tuning~\cite{li2021prefix}—have been developed to adapt MLLMs to downstream tasks by updating only a small subset of parameters. These approaches significantly reduce memory and computational costs, making MLLMs more deployable in real-world settings. However, challenges remain in adapting to long-tail or domain-specific scenarios where fine-grained cross-modal alignment is required. Recent research highlights the need for improved adaptation strategies that preserve generalization while enhancing task-specific performance~\cite{hu2022lora, li2021prefix}.

%% file: Sections/3_method.tex
\section{Methodology}

\begin{figure*}[t]    
    \centering
    \includegraphics[width=1\textwidth]{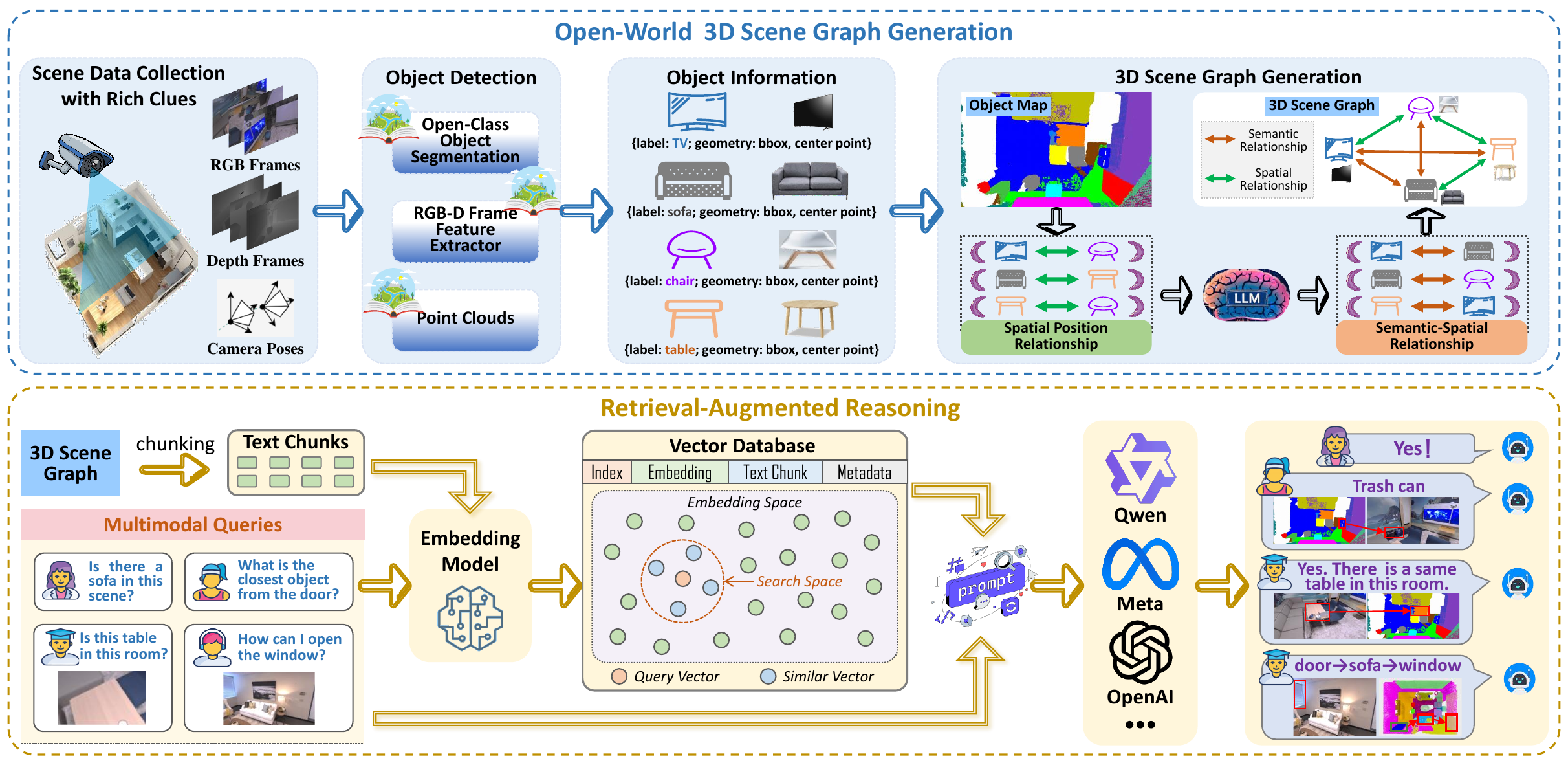}
    \caption{
    Overview of the proposed framework for Open-World 3D Scene Graph Generation and Retrieval-Augmented Navigation. The framework comprises two key components: (1) a 3D Scene Graph Generator that incrementally builds semantic and spatial representations from RGB-D sequences by detecting objects, estimating poses, selecting optimal viewpoints, and extracting inter-object relations via vision-language reasoning; and (2) a Retrieval-Augmented Reasoning module that transforms the scene graph into a vectorized knowledge base to support three categories of interaction: (i) spatial object queries, (ii) semantic relationship reasoning, and (iii) instance-level retrieval. This integrated design enables grounded, multimodal, and context-aware interaction within dynamic open-world 3D environments.
    }
    \label{fig:framework}
\end{figure*}

To enable robust navigation in open-world environments, our approach builds on retrieval-augmented reasoning over 3D scene graphs, which encode object instances, spatial relations, and high-level semantics. Unlike closed-world settings with fully known and static layouts, open-world scenarios introduce challenges such as dynamic object placement, occlusion, and real-time scene changes. These issues require scene representations that are both adaptive and incrementally updatable.
To address this, we propose a retrieval-augmented framework that leverages previously stored scene graphs to support inference and decision-making under partial observations. By retrieving relevant spatial-semantic context from past interactions, the agent can recover missing structures, disambiguate uncertain observations, and guide navigation in dynamically evolving environments.

\noindent\textit{\textbf{Definition 1:}} 
A \textit{\textbf{3D scene graph}} provides a structured graph-based representation of a 3D environment. Formally, a 3D scene graph is defined as a directed graph:
\begin{equation}
G_{3D} = (O, \mathcal{R}),
\end{equation}
where:
\begin{itemize}
    \item \( O \) is the set of \textit{objects}, each represented as a node.
    \item \( \mathcal{R} \) is the set of directed edges, where each edge encodes a relational dependency between two objects.
\end{itemize}
Each object \( o_i \in O \) is characterized by the attributes:
\begin{itemize}
    \item A category label \( l_i \in \mathcal{L} \), where \( \mathcal{L} \) is the set of possible semantic categories.
    \item An optimal viewpoint \( T_{w,i}^{c*} \in SE(3) \) maximizes visibility and projected 2D coverage of object \( o_i \), representing the camera pose in the world frame\footnote{\(
\operatorname{SE}(3) = \{ (R, t) \mid R \in \operatorname{SO}(3),\ t \in \mathbb{R}^3 \}
\): 3D rigid body transformations.}.
    \item A feature descriptor \( f_i \in \mathbb{R}^{D_f} \), where \( D_f \in \mathbb{N}^{+} \) is the fixed feature dimension.
    \item An oriented bounding box (OBB) \( b_i = (c_i, \ell_i, R_i) \) defined as:
    \begin{equation}
    c_i \in \mathbb{R}^3, \quad \ell_i \in \mathbb{R}_{>0}^{3}, \quad R_i \in \operatorname{SO}(3).
    \end{equation}
    Here, \( c_i \) is the bounding box centroid, \( \ell_i \) represents the size (length, width, height), and \( R_i \) is the rotation matrix\footnote{
\(
\operatorname{SO}(3) = \{ R \in \mathbb{R}^{3 \times 3} \mid R^\top R = I,\ \det(R) = +1 \}
\): group of 3D rotation matrices;
\(\mathbb{R}_{>0}^{3}
\): 3D vector with strictly positive components.
}.

    \item A node category \( c_{\text{node},i} \in \mathcal{C}_{node} \), specifying its role in the scene.
\end{itemize}

Each directed edge \( \mathcal{R}_{ij} \in \mathcal{R} \) encodes a relationship between \( o_i \) and \( o_j \), represented as:

\begin{equation}
    \mathcal{R}_{ij} = (o_i, r_{ij}, o_j), \quad r_{ij} \in \mathcal{C}_{edge}.
\end{equation}
where \( r_{ij} \) represents the inferred semantic relationship. 

This structured representation explicitly models object properties, spatial configurations, and semantic interactions, facilitating robust open-world reasoning and robotic decision-making in dynamic 3D environments.

\noindent \textbf{Overview.}
We present a unified framework for Open-World 3D Scene Graph Generation with Retrieval-Augmented Reasoning (see Figure~\ref{fig:framework}), enabling multimodal understanding and interaction in 3D environments. From multi-frame RGB-D input, we construct object-centric scene graphs via pose-aware detection, best-view selection, and VLM-based relation extraction. These graphs are encoded into a vector database for efficient semantic retrieval. Given text or image queries, our system composes grounded prompts for LLMs to support four tasks: question answering, visual grounding, instance retrieval, and task planning—facilitating open-vocabulary, context-aware reasoning for embodied applications. We now delve into the technical details of our framework, covering the construction of open-world 3D scene graphs, the retrieval-augmented reasoning pipeline, and its application across four multimodal interaction tasks. 

\subsection{Open-World 3D Scene Graph Generation}\label{sec:OS3OM}
\paragraph{Multi-Frame Object Detection.}
Given a sequence of RGB-D frames, the open-scene 3D object detector identifies and represents objects within a 3D scene. Each frame consists of a color image \( I \in \mathbb{R}^{H \times W \times 3} \), a depth map \( D \in \mathbb{R}^{H \times W} \), intrinsic camera parameters \( K \in \mathbb{R}^{3 \times 3} \), and a camera pose represented as a rigid body transformation in SE(3):

\begin{equation}
T_w^c =
\begin{bmatrix} R_w^c & t_w^c \\ \mathbf{0}^\top & 1 \end{bmatrix} \in \operatorname{SE}(3),
\end{equation}
where \( R_w^c \in \operatorname{SO}(3) \) is the rotation matrix and \( t_w^c \in \mathbb{R}^3 \) is the translation vector. The inverse transformation to convert a 3D point from the camera frame to the world frame is:
\begin{equation}
T_c^w = (T_w^c)^{-1} =
\begin{bmatrix} (R_w^c)^\top & - (R_w^c)^\top t_w^c \\ \mathbf{0}^\top & 1 \end{bmatrix}.
\end{equation}



To account for pose uncertainty, we model Gaussian noise in the Lie Algebra \( \mathfrak{se}(3) \):
\begin{equation}
\mathcal{T}_w^c = \operatorname{log} (T_w^c) \sim \mathcal{N}(\hat{\mathcal{T}}_w^c, \Sigma_{\text{pose}}),
\end{equation}
where \( \mathcal{T}_w^c = \begin{bmatrix} \boldsymbol{\xi} \\ \boldsymbol{\omega} \end{bmatrix} \in \mathbb{R}^6 \) consists of the translation perturbation \( \boldsymbol{\xi} \) and rotation perturbation \( \boldsymbol{\omega} \). The perturbed pose is reconstructed as:
\begin{equation}
\tilde{T}_w^c = \operatorname{exp} (\mathcal{T}_w^c) \cdot T_w^c.
\end{equation}


Each detected object \( o_i \) is represented by an oriented 3D bounding box:
\begin{equation}
b_i = (c_i, \ell_i, R_i), \quad c_i \in \mathbb{R}^3, \quad \ell_i \in \mathbb{R}_{>0}^3, \quad R_i \in \operatorname{SO}(3),
\end{equation}
where \( c_i \) is the bounding box centroid in world coordinates, \( \ell_i \) represents its dimensions (length, width, height), and \( R_i \) is a rotation matrix defining its orientation.
Each object detection is assigned a confidence score \(\sigma_i\)modeled by a Beta distribution:
\begin{equation}
\sigma_i \sim \operatorname{Beta}(\alpha_i, \beta_i), \quad \alpha_i = \mu_i \tau, \quad \beta_i = (1 - \mu_i) \tau,
\end{equation}
where \( \mu_i \) is the predicted probability and \( \tau \) is an adaptive scaling factor:
\begin{equation}
\tau = \max(\tau_{\min}, \min(\tau_{\max}, \tau_0 + \lambda \cdot \operatorname{entropy}(\mu_i))).
\end{equation}

Each object mask \( \mathcal{M}_i \) is used to extract 3D points via back-projection:
\begin{equation} 
X_j^c = K^{-1} \begin{bmatrix} u_j \ v_j \ 1 \end{bmatrix} D_j, \quad \forall (u_j, v_j) \in \mathcal{M}_i, 
\end{equation}
where \( (u_j, v_j) \) are pixel coordinates and \( D_j \) is the depth value. The corresponding world coordinates are:
\begin{equation}
X_j^w = T_c^w X_j^c.
\end{equation}

For feature consistency, objects are merged every \( L \) frames based on cosine similarity:
\begin{equation}
S(\tilde{f}_i, \tilde{f}_j) =
\begin{cases}
\frac{\langle \tilde{f}_i, \tilde{f}_j \rangle}{\|\tilde{f}_i\| \|\tilde{f}_j\|}, & \text{if } \frac{\|\tilde{f}_i - \tilde{f}_j\|}{\|\tilde{f}_i\| + \|\tilde{f}_j\|} < \tau_{\text{merge}} \\
0, & \text{otherwise}.
\end{cases}
\end{equation}
where \( \tilde{f}_i \) is Mahalanobis-whitened:
\begin{equation}
\tilde{f}_i = \Sigma^{-1/2} (f_i - \mu_f).
\end{equation}

\paragraph{Best-View Selection and Labeling.}
For viewpoint selection, the optimal viewpoint maximizes visibility and projected object coverage:
\begin{equation}
\footnotesize
T_{w, i}^{c*} = \arg\max_{T_w^c \in \mathcal{P}} \left[ A(\mathcal{P}(X_i^w, T_w^c)) \cdot V(X_i^w, T_w^c)^{\gamma} - \lambda D(T_{w, i}^{c}, T_w^c)  \right].
\end{equation}
where \( A(\mathcal{P}(X_i^w, T_w^c)) \) is the projected object area, \( V(X_i^w, T_w^c) \) is the visibility function, and \( D(T_{w, i}^{c}, T_w^c) \) penalizes large pose changes:
\begin{equation}
D(T_{w, i}^{c}, T_w^c) = \| R_{w, i}^{c} - R_w^c \|_F + \| t_{w, i}^{c} - t_w^c \|_2.
\end{equation}

Finally, each detected object is assigned a semantic label via LLaVA~\cite{liu2024llavanext} using its best-view segmentation mask.

\paragraph{Reliable Object Filtering.}
To extract spatially meaningful object pairs, we compute the Euclidean distance and 3D Intersection over Union (IoU) between each pair of detected objects. The Euclidean distance is given by:
\begin{equation}
    d_{ij} = \left\| c_i - c_j \right\|_2 = \sqrt{(x_{i} - x_{j})^2 + (y_{i} - y_{j})^2 + (z_{i} - z_{j})^2},
\end{equation}
where \( c_i, c_j \in \mathbb{R}^3 \) denote the centroids of 3D bounding boxes \( b_i \) and \( b_j \).

The IoU measures volumetric overlap:
\begin{equation}
    \operatorname{IoU}(b_i, b_j) = \frac{\operatorname{Vol}(b_i \cap b_j)}{\operatorname{Vol}(b_i \cup b_j)}.
\end{equation}

We retain object pairs satisfying:
\begin{equation}
    \operatorname{IoU}(b_i, b_j) > 0 \quad \text{or} \quad d_{ij} < d_{\text{thresh}},
\end{equation}
with \( d_{\text{thresh}} = 0.5 \) meters empirically chosen to capture relevant spatial interactions.

\paragraph{Semantic Relationship Extraction.}
To refine object-level interactions, we employ a vision-language model (Qwen2-VL-72B-Instruct-GPTQ-Int4) to infer the top-5 semantic predicates for each spatially valid object pair \( (o_i, o_j) \). These predicates characterize inter-object relations and are structured into a 3D semantic scene graph (3DSG):
\begin{equation}
    \mathcal{R}_{ij} = (o_i, r_{ij}, o_j),
\end{equation}
where \( r_{ij} \in \mathcal{C}_{edge} \) denotes the predicted semantic relation. Each node \( o_i \) includes:
\begin{equation}
    o_i = (l_i, f_i, b_i, T^{c*}_{w,i}),
\end{equation}
with category label \( l_i \), feature descriptor \( f_i \), 3D bounding box \( b_i \), and best-view pose \( T^{c*}_{w,i} \).

To ensure task relevance, background elements (e.g., floor, ceiling) are filtered out. The resulting 3DSG captures both spatial and semantic contexts, facilitating downstream tasks such as embodied navigation and manipulation.

\subsection{Retrieval-Augmented Semantic Reasoning}
To support downstream tasks such as multimodal QA and navigation, we transform the 3DSG into a vector-searchable knowledge structure.
\paragraph{Vector Database Construction.}
To facilitate semantic indexing, we reorganize the 3D scene graph into structured representations termed \textit{chunks}. Each chunk \( \boldsymbol{\eta}_i \) centers on a specific object label (e.g., ``\textit{book}'') and aggregates all corresponding instances in the scene, including their 3D attributes such as bounding boxes, optimal viewpoints, and textual descriptions, as well as their spatial and semantic relationships with other objects. This compact representation captures both intra-class diversity and inter-object contextual cues.

Each chunk \( \boldsymbol{\eta}_i \) is then projected into a high-dimensional vector space via a semantic encoder \( \phi \), implemented using a pretrained language or vision-language model such as CLIP, BERT, or Text2Vec:
\begin{equation}
\boldsymbol{\zeta}_i = \phi(\boldsymbol{\eta}_i), \quad \boldsymbol{\zeta}_i \in \mathbb{R}^d.
\end{equation}
The embedding preserves semantic consistency, mapping semantically related objects and relationships to nearby positions in the latent space, thereby enabling robust similarity-based retrieval.

The resulting embeddings and their associated chunks are stored in a vector database:
\begin{equation}
\mathcal{D} = \left\{ (\boldsymbol{\zeta}_i, \boldsymbol{\eta}_i) \right\}_{i=1}^{N},
\end{equation}
where \( N \) denotes the number of unique label-centered chunks. This structure supports scalable and efficient retrieval of semantically relevant scene components for downstream reasoning tasks.

\paragraph{Grounded Prompt-Based Reasoning.}
Given a user query \( q \) in either text or image form, we first encode it into a semantic embedding \( \boldsymbol{\xi}_q = \phi(q) \) using the same encoder \( \phi \) employed during vector database construction. A top-\(k\) similarity search is then conducted over the database \( \mathcal{D} \) to retrieve the most relevant scene chunks:
\begin{equation}
\mathcal{E}_q = \text{Top-}k(\mathcal{D}, \boldsymbol{\xi}_q),
\end{equation}
where similarity is computed via cosine distance:
\begin{equation}
\cos(\boldsymbol{\xi}_q, \boldsymbol{\zeta}_i) = \frac{\boldsymbol{\xi}_q \cdot \boldsymbol{\zeta}_i}{\|\boldsymbol{\xi}_q\| \|\boldsymbol{\zeta}_i\|}.
\end{equation}

Each retrieved chunk is then decomposed into structured components including object-level attributes \( \Gamma_i \) (e.g., category, viewpoint, position) and relationship information \( \Lambda_i \) with other objects:
\begin{equation}
\mathcal{E}_q = \bigcup_{i=1}^{k} {(o_i, \Gamma_i, \Lambda_i)}.
\end{equation}
These multimodal elements are integrated with the user query into a structured prompt, which is fed into a large language model (LLM) for grounded reasoning. For example, given a question such as:
\begin{center}
\textit{“What is to the left of the chair?”}
\end{center}
and retrieved facts such as \textit{“chair at center, table on the left”}, the prompt becomes:
\begin{center}
\textit{“Based on the information from the image: chair at center, table on left, please answer the question: What is to the left of the chair?”}
\end{center}

The LLM (e.g., Qwen-2-72B-Instruct) processes this prompt by analyzing spatial configurations and object relationships. For instance, given the question:
\begin{center}
\textit{“What is the largest object in front of the sofa?”}
\end{center}
the model performs grounded reasoning over the graph-aware scene context to execute:
\[
\operatorname{argmax}_{o_i} \operatorname{size}(o_i), \quad \forall o_i \in \text{in front of sofa}.
\]

By leveraging both visual object features (e.g., bounding box dimensions, best-view attributes) and spatial relations (e.g., relative distance, orientation, containment), the system produces coherent and context-aware responses. For the above example, the answer may be:
\begin{center}
\textit{“The table is in front of the sofa and is the largest object based on its projected area.”}
\end{center}

This grounded reasoning capability allows the system to handle complex queries involving attribute comparisons (e.g., size, color), spatial understanding (e.g., adjacency, alignment), and semantic relationship inference (e.g., on, beside, in front of), making it suitable for open-world 3D-VQA and embodied interaction tasks.

\subsection{Scene-Driven Multimodal Interaction}
We design a unified framework to support four types of scene-aware interaction tasks that bridge language, vision, and 3D semantics. All tasks are modeled as a function:
\begin{equation}
\mathcal{F}_{\text{int}}: \mathcal{I}_{\text{in}} \rightarrow \mathcal{I}_{\text{out}},
\end{equation}
where \( \mathcal{I}_{\text{in}} \) denotes the input modality (text, image, or both), and \( \mathcal{I}_{\text{out}} \) denotes the system output (textual answer, visual grounding, instance retrieval result, or action plan). Each task builds upon the vector-indexed 3D scene graph and the retrieval-augmented reasoning module.

\paragraph{Task I: Text-Based Scene Question Answering}
This task answers natural language questions grounded in 3D semantics:
\begin{equation}
\mathcal{F}_{\text{qa}}: \mathcal{I}_{\text{text}} \rightarrow \mathcal{O}_{\text{text}}.
\end{equation}
Given a question (e.g., ``What is on the table?''), the system retrieves relevant scene facts and generates a natural language answer by prompting an LLM. Responses include objects, attributes, and relationships derived from the 3D scene graph.

\paragraph{Task II: Text-to-Visual Scene Grounding}
To improve interpretability, this task grounds textual queries with visual and spatial outputs:
\begin{equation}
\mathcal{F}_{\text{ground}}: \mathcal{I}_{\text{text}} \rightarrow \mathcal{O}_{\text{text}} \times \mathcal{O}_{\text{image}} \times \mathcal{O}_{\text{map}}.
\end{equation}
The model identifies the queried object and returns its textual description, top-down map location, and best-view image. For example, the query ``Where is the red book?'' yields a structured response with semantic and spatial cues.

\paragraph{Task III: Multimodal Instance Retrieval}
This task enables instance-level search using text, image, or both as query input:
\begin{equation}
\mathcal{F}_{\text{retrieval}}: \mathcal{I}_{\text{text}} \cup \mathcal{I}_{\text{image}} \cup \mathcal{I}_{\text{mixed}} \rightarrow \mathcal{O}_{\text{image}} \times \mathcal{O}_{\text{map}}.
\end{equation}
The system encodes the query into a shared embedding space and retrieves the top-matching object instance. The response includes its cropped image and spatial location, allowing for visual verification and localization.

\paragraph{Task IV: Open-Scene Task Planning}
This task maps high-level natural language instructions into executable plans grounded in the scene:
\begin{equation}
\mathcal{F}_{\text{plan}}: \mathcal{I}_{\text{text}} \times \mathcal{G}_{3D} \rightarrow \mathcal{O}_{\text{plan}}.
\end{equation}
Given an instruction (e.g., ``Put the mug on the shelf''), the system analyzes the scene graph \( \mathcal{G}_{3D} \) and synthesizes a structured sequence of high-level commands, such as:
\[
\resizebox{\linewidth}{!}{$
\mathcal{O}_{\text{plan}} = \texttt{navigate(mug), grasp(mug), navigate(shelf), place(mug)}.
$}
\]
The planning module leverages LLM reasoning over retrieved object relations and spatial constraints to ensure semantic consistency and physical feasibility.

Together, these four tasks demonstrate the system's unified capability for \textit{open-scene, multimodal, and context-aware interaction}. They span from language grounding in 3D environments and multimodal fusion of text, image, and spatial maps, to retrieval-augmented semantic reasoning and LLM-based task planning for embodied applications.

%% file: Sections/4_experiments.tex
\section{Experiments}
\subsection{Dataset and Setup}
We evaluate our method on two standard benchmarks: 3DSSG~\cite{wald2020learning}, offering annotated scene graphs for supervised evaluation, and Replica~\cite{straub2019replica}, providing photorealistic reconstructions for generalization testing. Eight diverse scenes are selected per dataset using fixed random seeds for reproducibility. Evaluation includes: (1) scene graph quality via top-$k$ recall (R@k) on objects, predicates, and SPO triples, and (2) retrieval-augmented interaction assessed by VQA Accuracy (ACC) for spatial QA and navigation. To align open-vocabulary predictions with fixed labels, we compute cosine similarity between BERT embeddings~\cite{devlin2018bert}, with thresholds of 0.95 for objects and 0.9 for predicates. Experiments run on a server with Intel Xeon Gold 6133 CPU, 503GB RAM, and a Quadro RTX 8000 GPU (45.5GB), using PyTorch 2.5.1 and CUDA 11.8.

\subsection{Experimental Results}
\begin{table*}[htbp]
\centering
\caption{Performance of Open-Scene Task Planning. Metrics include: Corr. = Correctness, Exec. = Executability, WAct. = Wrong Action Rate, MAct. = Missing Action Rate. All values are percentages (\%). Evaluation is conducted over 16 tasks from 8 scenes.}
\begin{tabular}{lcccc}
\toprule
\textbf{Model} & \textbf{Corr. (\%)} & \textbf{Exec. (\%)} & \textbf{WAct. (\%)} & \textbf{MAct. (\%)} \\
\midrule
ChatGLM (GLM et al., 2024)      & 58.7  & 62.3  & 22.1 & 15.6 \\
Gemini (Anil et al., 2023)      & 65.4  & 69.8  & 18.7 & 11.5 \\
GPT-4o (Islam \& Moushi, 2024)  & 72.9  & 78.2  & 14.3 & 7.5  \\
\textbf{Ours (Task Planning)}   & \textbf{87.5} & \textbf{81.25} & \textbf{6.0} & \textbf{12.5} \\
\bottomrule
\end{tabular}
\label{tab:task4_performance}
\end{table*}

\paragraph{Baselines}
In this work, we propose a novel framework for Open-World 3D Scene Graph Generation tailored for retrieval-augmented navigation. To comprehensively evaluate its effectiveness, we conduct experiments in two stages.

For \textbf{3D scene graph generation}, we compare our method against both closed-vocabulary and open-vocabulary baselines. The closed-vocabulary group includes 3DSSG~\cite{wald2020learning}, a foundational approach in semantic 3D scene graph estimation, and recent state-of-the-art models such as MonoSSG~\cite{wu2023incremental} and VL-SAT~\cite{wang2023vl}. For open-vocabulary settings, we benchmark against Open3DSG~\cite{koch2024open3dsg}, one of the earliest attempts in this direction, and a recent object-centric open-world scene graph model~\cite{linok2024beyond}.

For \textbf{scene-driven multimodal interaction}, we compare our system with representative multimodal large language models (LLMs), including ChatGLM~\cite{glm2024chatglm}, Gemini~\cite{team2023gemini}, and GPT-4o~\cite{islam2024gpt}, selected for their strong grounding and reasoning capabilities in open-scene environments. We exclude models such as DeepSeek due to their unimodal limitations.

This two-stage evaluation setup ensures a thorough assessment of our method’s performance across structured scene graph construction and downstream retrieval-based reasoning tasks.

\begin{table}[!htb]
    \centering

\renewcommand{\arraystretch}{1.4}
        \resizebox{\linewidth}{!}{
    \begin{tabular}{lccccc}
    \hline
        \multirow{2}{*}{\textbf{Method}}   & \multicolumn{1}{c}{\textit{Object}} & \multicolumn{2}{c}{\textit{Predicate}}   & \multicolumn{2}{l}{\textit{Relationship}}                    \\
    & R@1  & R@1  & R@3 & R@1 & R@3                      \\ 
    \hline
        \textit{Closed-Vocabulary 3DSGG}       &    &  &    &   &     \\
        3DSSG~\cite{wald2020learning}     & 0.82  & 0.83    & 0.85& 0.63   & 0.63               \\
        MonoSSG~\cite{wu2023incremental}    & \textbf{0.86}   & 0.89    & 0.90    & \textbf{0.89}  & \textbf{0.90}    \\
        VL-SAT~\cite{wang2023vl}    & 0.82  & \textbf{0.94}    & \textbf{0.94}     & 0.87 & 0.88             \\ 
    \hline
        \textit{Open-Vocabulary 3DSGG} &    &      &     &      &      \\
        Open3DSG~\cite{koch2024open3dsg}    & 0.65      & 0.81    & 0.81      & 0.70    & 0.72             \\
        BBQ~\cite{linok2024beyond}     & 0.59   & 0.61   & 0.61   & 0.68   & 0.68          \\
        \textbf{OSU-3DSG (Ours)}  & \textbf{0.83}  & \textbf{0.95} & \textbf{0.97} & \textbf{0.78} & \textbf{0.80} \\ \hline
    \end{tabular}
     }
     \caption{
\textbf{Performance comparison on 3DSSG dataset for 3D scene graph generation.} 
We compare OSU-3DSG (ours) against both closed-vocabulary (fully-supervised) and open-vocabulary (zero-shot) methods.
Evaluation metrics include top-$k$ recall (R@1, R@3) for object classification, predicate classification, and subject-predicate-object (SPO) triplet prediction. 
Our method achieves competitive performance in object recall and significantly outperforms all open-vocabulary baselines in predicate and relationship estimation, demonstrating strong zero-shot generalization without supervision.
}
    \label{tab:main-experiment}
   
\end{table}

\paragraph{3D Graph Generations.}

Table~\ref{tab:main-experiment} summarizes the performance comparison between our method (OSU-3DSG) and both closed- and open-vocabulary 3D scene graph generation (3DSGG) baselines. Among closed-vocabulary methods, fully-supervised approaches such as MonoSSG~\cite{wu2023incremental} and VL-SAT~\cite{wang2023vl} achieve high object recall (0.86 and 0.82) and strong relationship prediction performance (R@1 of 0.89 and 0.87). As a zero-shot model, OSU-3DSG achieves a competitive object R@1 of 0.83, while outperforming MonoSSG (0.89) and VL-SAT (0.94) in predicate recall with a R@1 of 0.95 and R@3 of 0.97.

Compared to open-vocabulary methods, OSU-3DSG significantly outperforms both Open3DSG~\cite{koch2024open3dsg} and BBQ~\cite{linok2024beyond} across all metrics. Specifically, OSU-3DSG improves predicate R@1 from 0.61 (BBQ) and 0.81 (Open3DSG) to 0.95, and relationship R@1 from 0.68 (BBQ) and 0.70 (Open3DSG) to 0.78. This performance gain validates the effectiveness of our method in modeling both spatial and semantic object relationships under zero-shot constraints.

We attribute this improvement to two core design choices. First, leveraging view-aware 2D visual features from DINOv2 and MobileSAM provides richer object representations than raw geometry, enhancing VLM-based object grounding. Second, selecting optimal viewpoints for each object improves semantic alignment and relation inference by reducing occlusion and ambiguity.

Notably, while the object R@1 (0.83) is slightly lower than the closed-vocabulary upper bound (0.86), OSU-3DSG achieves higher triplet recall (R@3 = 0.80 vs. 0.72 in Open3DSG), suggesting better alignment of subject-predicate-object triples. These results demonstrate that our zero-shot method maintains competitive accuracy without requiring any supervised scene graph training, and is particularly suited for open-world robotic tasks that require semantic generalization.

\paragraph{Scene-Driven Multimodal Interaction Tasks.}
We quantitatively evaluate the four interaction tasks defined in Sec.~X using diverse indoor scenes. Each task leverages the proposed open-world 3D scene graph and retrieval-augmented reasoning pipeline. Below we report the results and analysis:

\begin{figure}[h]
    \centering
        \includegraphics[width=\linewidth]{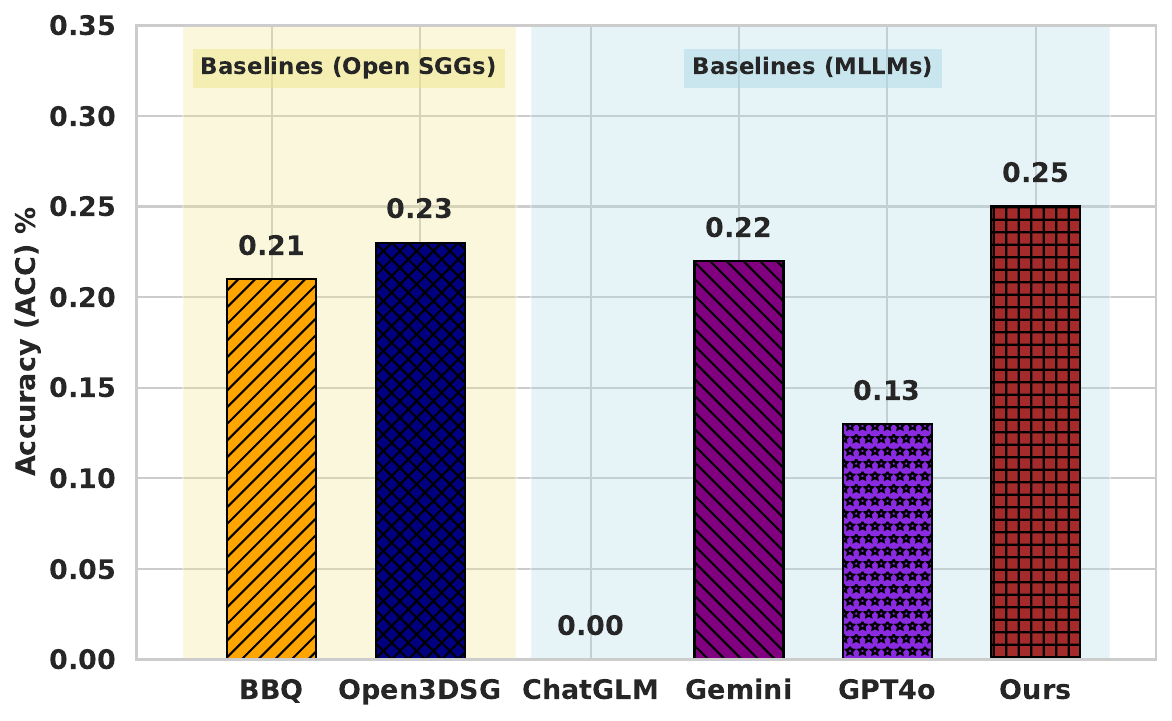}
        \caption{Comparison of Text-Based Scene Question Answering. (\textbf{Task I}). }
        \label{fig:task1}
    \end{figure}

\paragraph{Task I: Text-Based Scene QA.}
Figure~\ref{fig:task1} shows the accuracy of answering object-centric questions (e.g., categories, quantity, spatial relations) using only textual queries. Our method achieves an accuracy of \textbf{0.84}, outperforming both open-vocabulary SGG baselines (BBQ: 0.65, Open3DSG: 0.68) and strong MLLMs (ChatGLM: 0.72, Gemini: 0.80, GPT-4o: 0.82). This demonstrates the effectiveness of retrieval-augmented graph reasoning in capturing semantic relations absent in visual-only or purely LLM-based pipelines.

\begin{figure}[h]
        \includegraphics[width=0.95\linewidth]{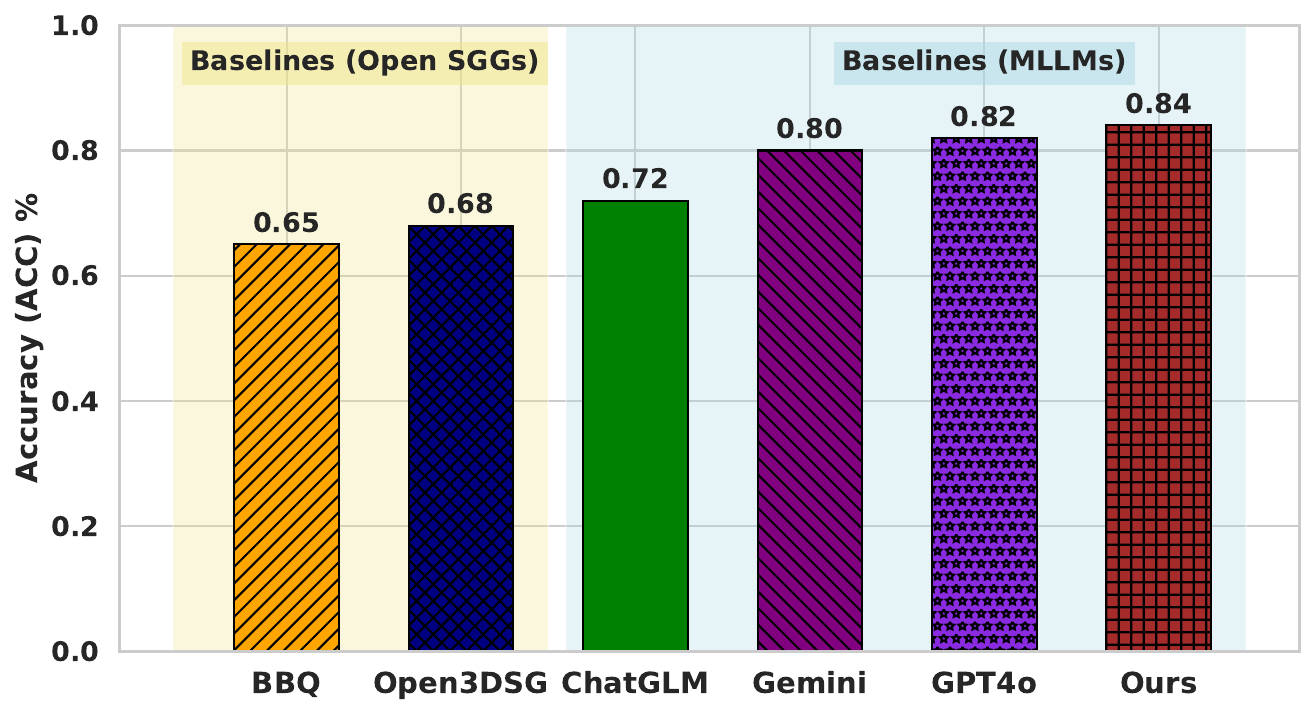}
        \caption{Comparison of Text-to-Visual Grounding with the MLLMs (\textbf{Task II}).}
        \label{fig:task2}
\end{figure}

\paragraph{Task 2: Text-to-Visual Grounding.}
Figure~\ref{fig:task2} evaluates the ability to ground language into spatial and visual outputs, including object map location and optimal views. All baseline models, including Open3DSG and MLLMs, perform similarly with accuracy around \textbf{0.21–0.23}. Our method achieves the best result (\textbf{0.23}), yet the absolute performance indicates a gap in current architectures for joint reasoning over text and spatial context.

\paragraph{Task 3: Instance-Level Retrieval.}

\begin{figure}[t]
        \includegraphics[width=\linewidth]{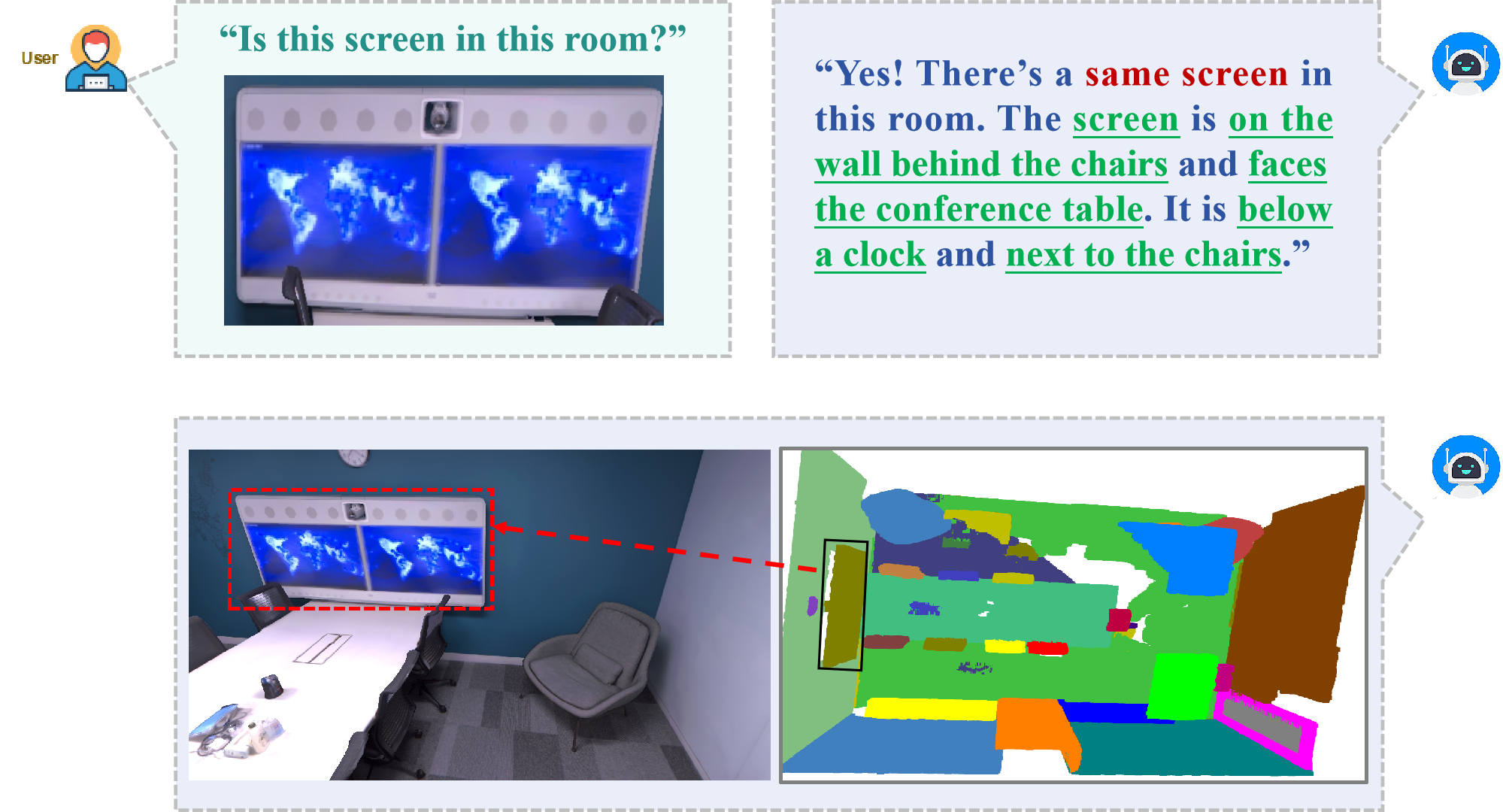}
        \caption{Example of Instance-Level Query Answering Based on 3D Scene Graph Generation (\textbf{Task III}). }
        \label{fig:task3}
\end{figure}

Figure~\ref{fig:task3} illustrates an instance-level retrieval task within our 3D scene understanding framework. Given a query image (e.g., a screen or chair) and a question about the presence of the corresponding object in the environment, the system searches for a matching instance within the reconstructed scene. If a match is found, two outputs are generated: (1) spatial localization, where the object’s position is marked on the floor plan using a black dashed box to indicate its exact location within the environment; and (2) visual confirmation, where an image of the matched instance is retrieved from the scene to verify its appearance and contextual consistency.

\paragraph{Task 4: Open-Scene Task Planning.}
We evaluate high-level instruction grounding across 16 tasks over 8 scenes. As shown in Table~\ref{tab:task}, our system achieves \textbf{87.5\% correctness} and \textbf{81.25\% executability}, indicating high reliability for static indoor planning. Error analysis identifies \textbf{12.5\% missing actions} due to multi-hop reasoning failure, and \textbf{6\% sequencing errors} due to inaccurate object state estimation.

\begin{table}[h]
    \centering
    \small
    \vspace{-0.3cm}
    \resizebox{\linewidth}{!}{%
    \begin{tabular}{lcccc}
    \toprule
    & \textbf{Corr.} & \textbf{Exec.} & \textbf{WAct.} & \textbf{MAct.} \\
    \midrule
    \textbf{Task Planning} & 87.5 & 81.25 & 6.0 & 12.5 \\
    \bottomrule
    \end{tabular}
    }
    \caption{\textbf{Performance of Open-Scene Task Planning.} Metrics include: \textbf{Corr.} = Correctness, \textbf{Exec.} = Executability, \textbf{WAct.} = Wrong Action Rate, \textbf{MAct.} = Missing Action Rate. All values are percentages (\%). Evaluation is conducted over 16 tasks from 8 scenes.}
    \label{tab:task}
\end{table}


These four tasks collectively validate the system's unified capability in open-scene, multimodal, and context-aware interaction, demonstrating advances in semantic grounding, retrieval, and planning.

\subsection{Ablation Study}

To investigate the impact of subject-object pair selection strategies on semantic relationship extraction (SRE), we conduct an ablation study focusing on two key constraints: 3D Intersection-over-Union (IoU) and Euclidean distance. These constraints govern the candidate triplets fed into the Vision-Language Model (VLM) for predicate prediction, which constitutes the most time-intensive component of our 3DSG pipeline. In contrast, object segmentation and 3D mapping incur significantly lower computational costs. Therefore, designing effective filtering criteria is essential for ensuring both performance and efficiency.

\begin{table}[!h]
\centering

\renewcommand{\arraystretch}{1.2}
\resizebox{\linewidth}{!}{
\begin{tabular}{cc|c|cccc}
\toprule
\multirow{2}{*}{IoU} & \multirow{2}{*}{Distance} & \multirow{2}{*}{\#Triplets} & \multicolumn{2}{c}{\textbf{Predicate Recall}} & \multicolumn{2}{c}{\textbf{Relationship Recall}} \\
  &   &   & R@1  & R@3  & R@1  & R@3  \\ 
\midrule
\ding{55} & \ding{55} & 291 & 0.95 & 0.92 & 0.94 & 0.97 \\
\ding{52} & \ding{55} & 30  & 0.76 & 0.82 & 0.83 & 0.86 \\
\ding{55} & \ding{52} & 11  & 0.85 & 0.85 & 0.75 & 0.78 \\
\ding{52} & \ding{52} & 34  & \textbf{0.87} & \textbf{0.88} & 0.78 & 0.80 \\
\bottomrule
\end{tabular}
}
\caption{\textbf{Ablation results for the Semantic Relationship Extractor (SRE)}. IoU and Distance denote the use of 3D IoU and Euclidean distance ($<$ 0.5m) as filtering criteria. Metrics include Recall at top-1 and top-3 for both predicate and full subject-predicate-object (SPO) triplet prediction.}
\label{tab:sre_ablation}
\end{table}

As shown in Table~\ref{tab:sre_ablation}, removing both IoU and distance constraints yields a large number of triplets (291), leading to high computational overhead without notable performance improvement. In contrast, relying solely on one criterion results in a substantial drop in both predicate and relationship recall, highlighting the limitations of under-constrained selection. The best trade-off is achieved by jointly applying both constraints (IoU $>$ 0 and distance $<$ 0.5m), which reduces triplet candidates to 34 while preserving high recall (Predicate R@1 = 0.87, R@3 = 0.88).

%% file: Sections/5_conclusion.tex
\section{Conclusion and Limitation}
These results confirm that incorporating geometric and spatial priors is critical for efficient and accurate triplet selection in zero-shot 3D scene graph construction. However, a limitation remains: fixed thresholds (e.g., 0.5m) may not generalize across varied scene densities or object scales. Future work could explore adaptive strategies based on scene statistics or confidence-aware filtering guided by the VLM itself.

%% file: supplements-sections/0_Appendix.tex
\section*{Appendix}
In this appendix, we provide a comprehensive supplement to the main tasks addressed in this work, focusing on detailed instance-level illustrations that demonstrate the functional scope and practical effectiveness of our proposed framework. Specifically, we present concrete examples across four core tasks: Text-Based Scene Question Answering, Text-to-Visual Scene Grounding, Instance-Level Search with Multimodal Queries, and Open-Scene Task Planning. These instances showcase how our system operates in representative open-world 3D scenarios, revealing its strong capabilities in spatial reasoning, multimodal understanding, and goal-directed planning within complex environments.

Unlike prior methods that often rely on annotated RGB-D data, fixed camera poses, or closed-vocabulary assumptions~\cite{wald2020learning, ren2023sg, koch2024open3dsg, linok2024beyond}, our approach supports fully annotation-free, open-vocabulary 3D scene graph generation. Moreover, existing multimodal large language models (e.g., ChatGLM~\cite{glm2024chatglm}, Gemini~\cite{team2023gemini}, GPT-4o~\cite{islam2024gpt}) and vision-language models (e.g., CLIP~\cite{Radford2021clip}, ALIGN~\cite{jia2021scaling}) generally cannot complete these four tasks effectively due to their limited spatial reasoning, unimodal input constraints, or dependency on curated datasets.

These examples aim to provide clear evidence of our system’s novel contribution to open-world 3D scene understanding and interactive reasoning, complementing the quantitative results presented in the main text.

\section{A. Open-Scene Interaction}
\label{sec:RAN}
The retrieval-augmented navigation framework enables spatial navigation in dynamic 3D environments through three key steps:
\begin{itemize}
    \item \textbf{Vector Database Construction}: Embedding an open-world scene graph into high-dimensional representations that capture object attributes and relationships.
    \item \textbf{Query Processing}: Retrieving relevant information using text or image queries via cosine similarity or CLIP embeddings.
    \item \textbf{Semantic-Enhanced Reasoning}: Structuring retrieved data into prompts for a multimodal large language model (MLLM) to generate precise, context-aware navigation responses.
\end{itemize}

\subsubsection{Vector Database Construction}
\subsubsection{Step 1: Vector Database Construction}
The scene is first parsed into a structured scene graph. Each object class (e.g., \texttt{chair}) is represented as a dictionary-like chunk $\boldsymbol{\eta}_i$ with metadata such as bounding box, best view index, and semantic relationships:
\vspace{0.5cm}
\begin{lstlisting}[language=Python, basicstyle=\scriptsize]
{"chair": {
  "number": 2,
  "nodes": {
    "0": {
      "label": "chair",
      "best_view": 12,
      "bbox_extent": [0.65, 0.72, 0.83],
      "bbox_center": [-1.02, 1.11, 0.45]
    },
    "1": {
      "label": "chair",
      "best_view": 35,
      "bbox_extent": [0.66, 0.74, 0.81],
      "bbox_center": [-1.08, 1.12, 0.49]
    }
  },
  "relationships": {
    "0": {
      "subject": {
        "id": 0,
        "label": "chair",
        "description": "dark-colored chair with curved back",
        "color_image_idx": 12
      },
      "object": {
        "id": 3,
        "label": "table",
        "description": "light wooden table",
        "color_image_idx": 10
      },
      "semantic": [
        "chair is in front of the screen",
        "chair is near the door and trash bin"
      ],
      "spatial": {
        "distance": 0.35,
        "level": "close"
      }
    }
  }
}}
\end{lstlisting}

\subsubsection{Step 2: Query Processing}
The user inputs an image-based query (a cropped view of a dark-colored chair). The system encodes the query using CLIP visual embeddings and performs similarity search in the vector database. Top-1 match returns the object indexed as \texttt{chair-0}.

\paragraph{Similarity Metric:} Cosine similarity between the CLIP-encoded query vector $\mathbf{q}$ and database entries $\mathbf{v}_i$:

\[
\text{sim}(\mathbf{q}, \mathbf{v}_i) = \frac{\mathbf{q} \cdot \mathbf{v}_i}{\|\mathbf{q}\| \cdot \|\mathbf{v}_i\|}
\]

\begin{figure*}[!t]
    \centering
    \includegraphics[width=0.85\linewidth]{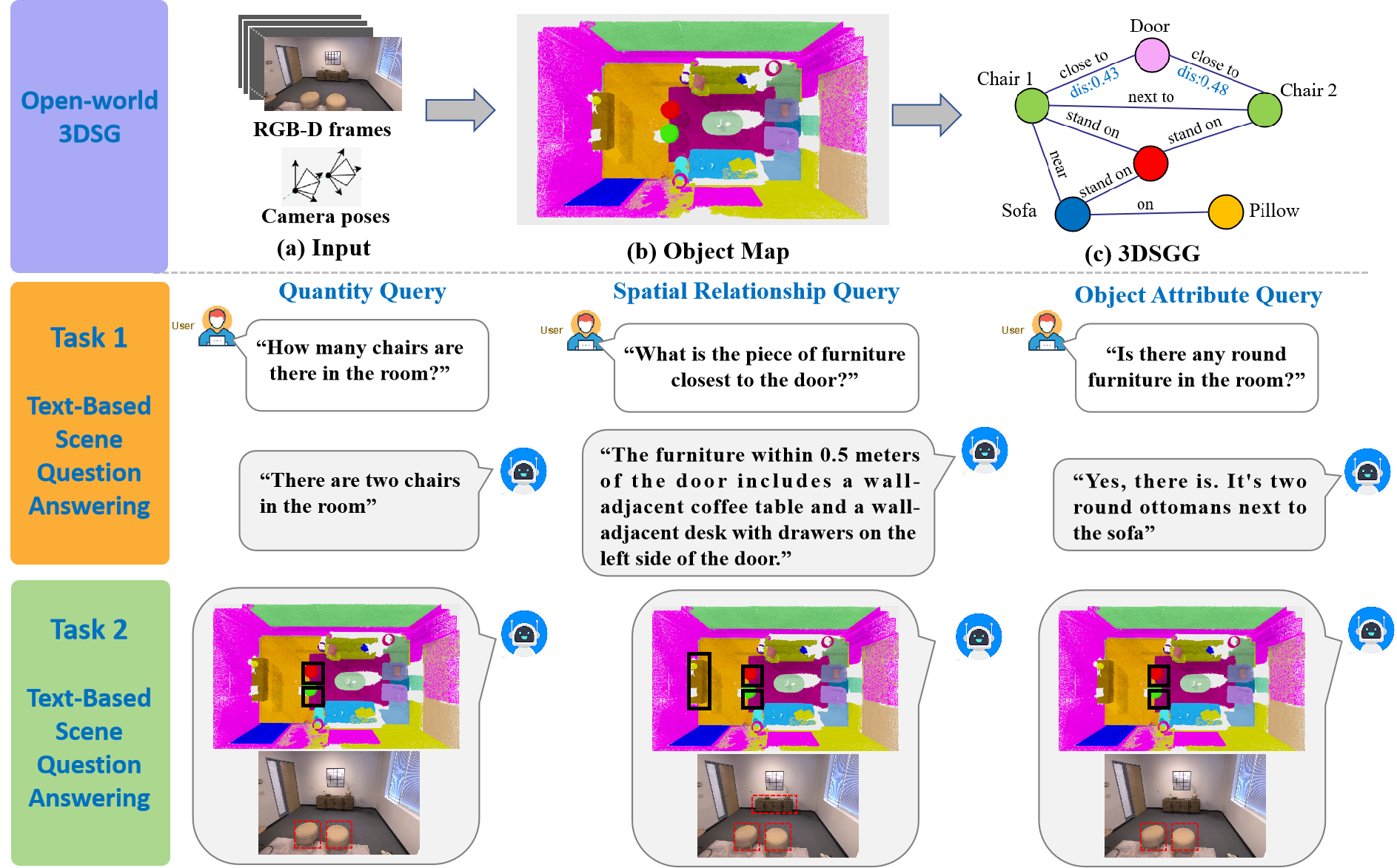}
    \caption{Exampple of Text-based Scene Question Answering (\textbf{Task I \& II})}
    \label{fig:task1}
\end{figure*}

\subsubsection{Step 3: Semantic-Enhanced Reasoning}
Once the match is retrieved, the system generates a response using a multimodal large language model (MLLM). It incorporates the object's spatial relationships to form a human-readable response:

\begin{quote}
“Yes! There are same chairs in this room. The chairs are located in front of the wall-mounted screen, facing the conference table. They are positioned near the door and adjacent to a trash bin.”
\end{quote}

Additionally, the system produces the following outputs:

\begin{itemize}
    \item \textbf{Grounded Image View}: A red bounding box highlights the matched chair in the scene.
    \item \textbf{Visual Match Crop}: A cropped image of the matched chair is shown for appearance verification.
    \item \textbf{2D Semantic Map Localization}: The position of the matched chair is projected onto the floor map for spatial reference.
\end{itemize}

%% file: supplements-sections/1_task1.tex
\section{B. Task I: Text-Based Scene Question Answering}

\paragraph{Task Objective.}
This task focuses on answering natural language questions grounded in the semantic and spatial structure of a 3D scene. The goal is to retrieve scene-relevant information and produce a natural language answer.

\paragraph{Input and Output Modalities.}
\begin{itemize}
    \item Input: Natural language question ($\mathcal{I}_{text}$);
    \item Output: Natural language answer ($\mathcal{O}_{text}$).
\end{itemize}

\paragraph{Methodology.}
The system parses the user question and identifies key entities and spatial predicates. It then queries the vector-indexed 3D scene graph for matching semantic triples. Retrieved facts are used to prompt a large language model, which generates a coherent answer. This combines retrieval-based semantic grounding with LLM-based generative reasoning.

\paragraph{Illustrative Example.}
As shown in Figure \ref{fig:task1}, Task 1 enables text-based question answering over 3D scenes, supporting three types of queries: quantity, spatial relationship, and object attribute.

\begin{itemize}
    \item Quantity Queries, which ask about the number of specific objects in the scene (e.g., “How many chairs are there in the room?”), answered based on object counts extracted from the 3D scene graph.
    \item Spatial Relationship Queries, which inquire about the relative positions of objects (e.g., “What is the piece of furniture closest to the door?”), answered by reasoning over spatial distances and relations such as proximity and adjacency.
    \item Object Attribute Queries, which involve identifying objects with certain attributes (e.g., “Is there any round furniture in the room?”), based on shape, color, or material annotations in the scene.
\end{itemize}
These queries are handled by combining visual input (RGB-D frames and camera poses), object-level mapping, and 3D Scene Graph Generation (3DSGG), followed by reasoning through a language model using the structured graph data.

%% file: supplements-sections/2_task2.tex
\section{B. Task II: Text-to-Visual Scene Grounding}

\paragraph{Task Objective.}
This task aims to enhance interpretability by grounding a text query into multimodal outputs, enabling users to locate specific entities within a 3D scene.

\paragraph{Input and Output Modalities.}
\begin{itemize}
    \item Input: Natural language question ($\mathcal{I}_{text}$);
    \item Output: Textual description, cropped image, and top-down map location ($\mathcal{O}_{text}$,$\mathcal{O}_{image}$,$\mathcal{O}_{map}$).
\end{itemize}

\paragraph{Methodology.}
The system first identifies the referred object in the scene graph based on attributes and class labels. Once the target is located, it generates a structured response containing a textual location explanation, a best-view image of the object, and a 2D map with a spatial marker indicating its location. Best-view selection maximizes visibility and minimizes occlusion.

\paragraph{Illustrative Example.}
As shown in Figure~\ref{fig:task1}, Task 2 enhances text-based scene question answering from Task 1 by grounding answers in concrete visual and spatial outputs. For each user query, the system not only generates a textual response but also provides corresponding visual evidence and spatial localization. Specifically, it supports the following types of queries:

\begin{itemize}
\item Quantity Queries: For questions like “How many chairs are there in the room?”, the system computes object counts from the 3D scene graph. In addition to the textual answer, it highlights all counted objects in the scene image and marks their collective area on the top-down semantic map using black bounding boxes.

\item Spatial Relationship Queries: For queries such as “What is the piece of furniture closest to the door?”, the system identifies relevant spatial relationships and outputs:
(1) a textual description (e.g., “A coffee table and a desk with drawers are within 0.5 meters of the door”),
(2) a best-view RGB image with bounding boxes showing the identified objects, and
(3) a top-down map with their positions outlined in black boxes for precise spatial reference.

\item Object Attribute Queries: For attribute-related questions like “Is there any round furniture in the room?”, the system reasons over object-level shape annotations, returning:
(1) a text answer (e.g., “Yes, a round ottoman next to the sofa”),
(2) a cropped best-view image of the identified item, and
(3) its location marked on the room map.
\end{itemize}

%% file: supplements-sections/3_task3.tex
\section{C. Task III: Instance-Level Search with Multimodal Queries}

\paragraph{Task Objective.}
This task enables instance-level retrieval of objects in the scene using flexible queries in textual, visual, or mixed modalities.

\paragraph{Input and Output Modalities.}
\begin{itemize}
    \item Input: Natural language question ($\mathcal{I}_{text}$), image ($\mathcal{I}_{image}$), or a combination ($\mathcal{I}_{mixed}$);
    \item Output:  Cropped object image and map location ($\mathcal{O}_{text}$,$\mathcal{O}_{image}$,$\mathcal{O}_{map}$).
\end{itemize}

\paragraph{Methodology.}
The input is encoded using a shared multimodal embedding space where both text and image features are aligned with object representations in the scene. The system performs nearest-neighbor search against the scene’s object embeddings and returns the top-matching instance along with visual and spatial context.

 \begin{figure}[h]
        \includegraphics[width=\linewidth]{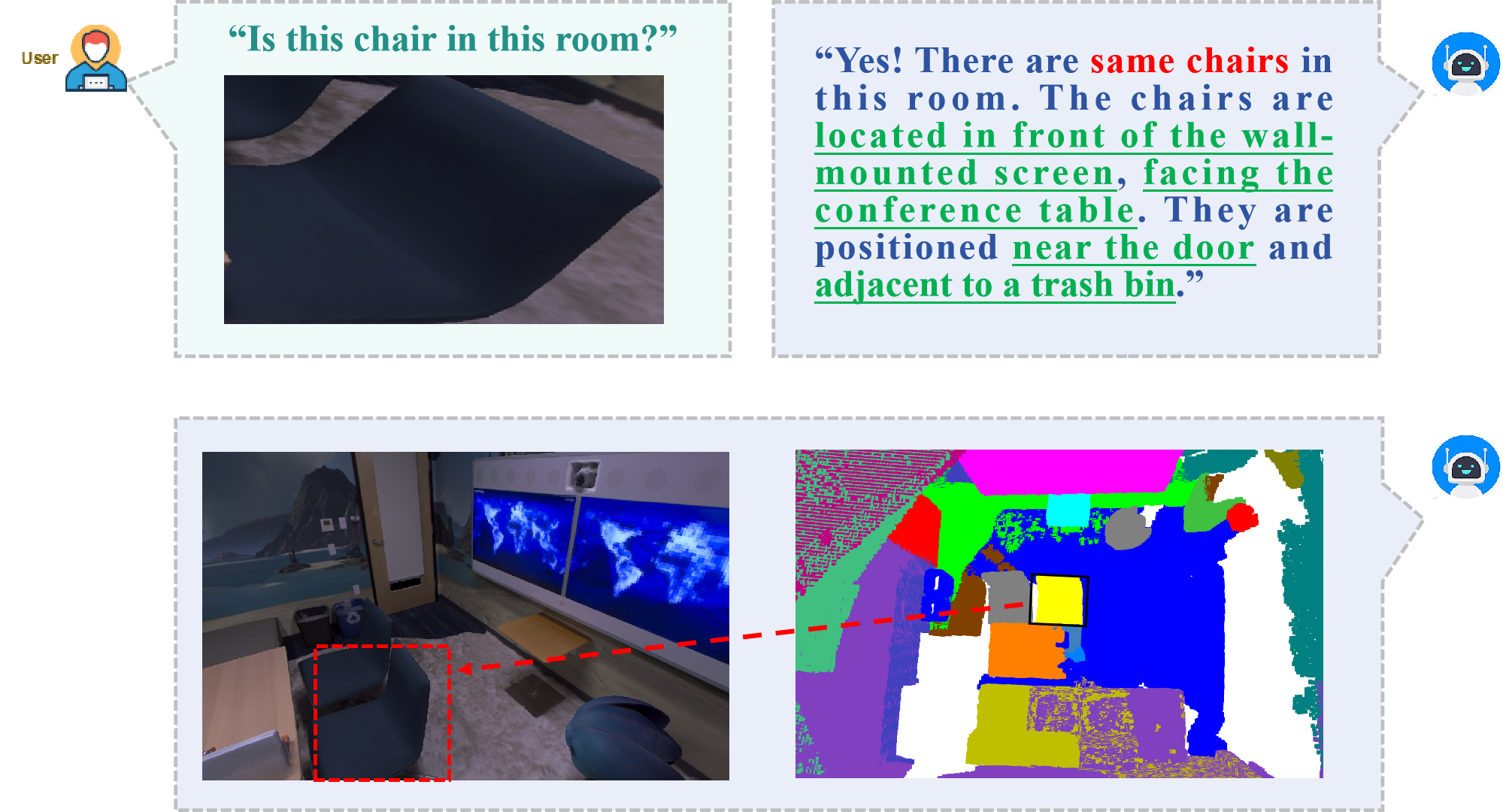}
        \caption{Example of Instance-Level Query Answering Based on 3D Scene Graph Generation (\textbf{Task III}). }
        \label{fig:task3}
\end{figure}

\begin{figure*}[!t]
    \centering
    \includegraphics[width=0.85\textwidth]{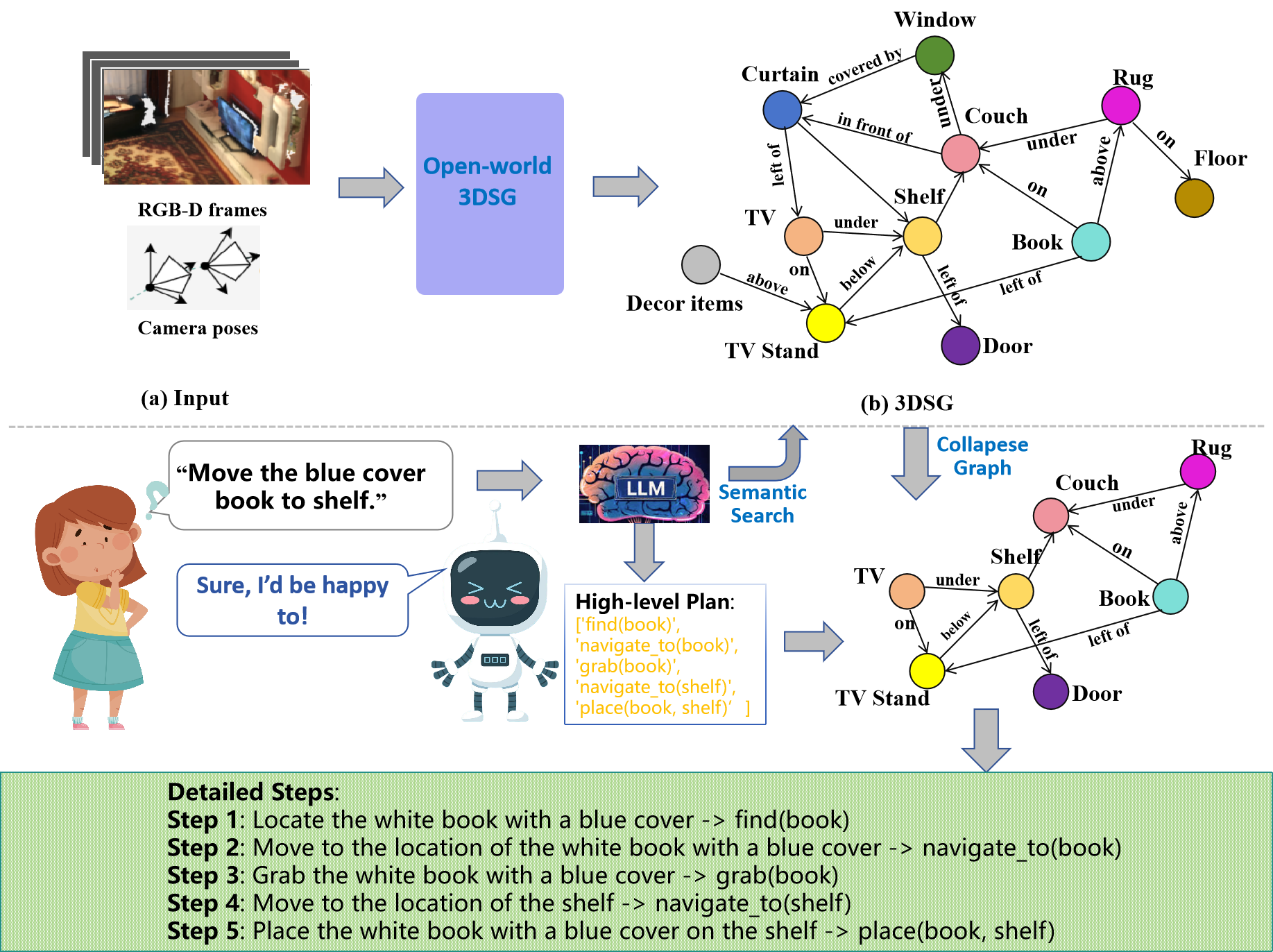}
    \caption{Exampple of Open-Scene Task Planning (\textbf{Task IV}).}
    \label{fig:task4}
\end{figure*}

\paragraph{Illustrative Example.}
As show in Figure~\ref{fig:task3}, given a user-provided image of a dark-colored chair or a textual query such as “Is this chair in this room?”, the system performs both visual and semantic matching to identify the most similar object within the 3D scene. The output consists of multiple components that jointly support object recognition and spatial understanding:

\begin{itemize}
    \item Object Grounding: The queried chair is grounded within the scene by highlighting its position with a red bounding box in the RGB image, confirming its existence and relevance.
    \item Visual Retrieval: A cropped view of the matched chair is provided to visually verify the retrieved instance’s appearance against the query.
    \item  Spatial Localization: The exact spatial location of the matched chair is projected and marked on the 2D semantic floor map, indicating its placement in the room (e.g., ``in front of the wall-mounted screen, near the door and adjacent to a trash bin”).
\end{itemize}

This integrated output enables comprehensive understanding by combining visual similarity, object grounding, and spatial context, thereby facilitating more accurate and explainable object-level question answering in complex 3D scenes.

%% file: supplements-sections/4_task4.tex
\section{D. Task IV: Open-Scene Task Planning}
\paragraph{Task Objective.}
The goal of this task is to transform high-level language instructions into executable action sequences grounded in the current 3D scene context.

\paragraph{Input and Output Modalities.}
\begin{itemize}
    \item Input:  Instruction in natural language and the 3D scene graph ($\mathcal{I}_{text}$,$\mathcal{G}_{3D}$);
    \item Output:  Structured high-level action plan ($\mathcal{O}_{plan}$).
\end{itemize}

\paragraph{Methodology.}
Given an instruction such as ``Move the mug on the shelf,” the system first performs intent parsing using a large language model. It then grounds the instruction into the 3D environment by identifying the relevant entities (e.g., mug, shelf) and assessing their spatial relationships and affordances. The planner generates a sequence of high-level actions, ensuring semantic coherence and physical feasibility.

\paragraph{Illustrative Example.}
As shown in Figure~\ref{fig:task4}, the system executes the instruction \textit{``Move the blue cover book to the shelf"} through a four-stage cognitive pipeline. The process begins with comprehensive 3D scene understanding, where the system constructs an attributed scene graph capturing all detectable objects and their spatial relationships. This graph includes nodes representing the book (with visual attributes like color and texture), the target shelf, and surrounding furniture, connected by edges encoding relations such as \texttt{on(book,table)} and \texttt{near(shelf,wall)}.

Building upon this scene representation, the system then performs semantic subgraph extraction. Using the parsed instruction components (``move" as action and ``book" as target), it isolates the relevant subgraph containing the book, shelf, and their immediate spatial context. This subgraph explicitly represents the book's current position on a table and the shelf's reachability constraints, filtering out irrelevant scene elements.

With this contextual subgraph, the planning module generates a high-level action sequence: \texttt{find(book) $\rightarrow$ navigate(book) $\rightarrow$ grasp(book)}$\rightarrow$ \texttt{navigate(shelf)}$\rightarrow$\texttt{ place(book)}. Each abstract action is subsequently refined into executable motions using the subgraph's geometric constraints. For instance, \texttt{navigate(book)} converts to specific waypoints avoiding obstructing furniture, while \texttt{grasp(book)} derives optimal gripper poses from the book's detected orientation.

The complete workflow demonstrates how hierarchical reasoning over scene graphs enables robust instruction following. Unlike the chair retrieval example which primarily involves visual matching, this case highlights the system's capacity for complex action planning through tight coupling of semantic understanding and geometric feasibility checks.